\documentclass{article}

\usepackage{arxiv}

\usepackage[utf8]{inputenc} 
\usepackage[T1]{fontenc}    
\usepackage{hyperref}       
\usepackage{url}            
\usepackage{booktabs}       
\usepackage{amsfonts}       
\usepackage{nicefrac}       
\usepackage{microtype}      
\usepackage{graphicx}
\usepackage{natbib}
\usepackage{doi}
\usepackage{framed}

\usepackage{caption}
\usepackage{subcaption}
\usepackage{array,tikz}
\usetikzlibrary{tikzmark,arrows.meta}
\newcolumntype{M}{>{\centering\mbox{}\vfill\arraybackslash}m{50pt}<{\vfill}}
\usepackage{tabularx}

\usepackage{amsthm,thmtools,mathtools,bm,cancel}
\theoremstyle{plain}

\theoremstyle{definition}
\newtheorem{defn}{Definition}[section]

\newtheorem{spec}{Specification}[section]

\theoremstyle{remark}

\usepackage{tipa}

\usepackage[acronym]{glossaries}
\newacronym{covid19}{COVID-19}{Coronavirus Disease 2019}

\title{Linguistic inspired graph analysis}

\author{ \href{https://orcid.org/0000-0003-0338-4567}{\includegraphics[scale=0.06]{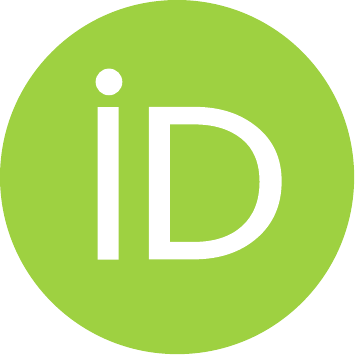}\hspace{1mm}Andrew Broekman}\\
	Department of Computer Science\\
	University of Pretoria\\
	Pretoria, South Africa\\
	\texttt{andrew.broekman@up.ac.za} \\
	\And
	\href{https://orcid.org/0000-0002-5270-243X}{\includegraphics[scale=0.06]{orcid.pdf}\hspace{1mm}Linda Marshall} \\
	Department of Computer Science\\
	University of Pretoria\\
	Pretoria, South Africa\\
	\texttt{lmarshall@cs.up.ac.za} \\
}

\renewcommand{\headeright}{Technical Report}
\renewcommand{\undertitle}{Technical Report}
\renewcommand{\shorttitle}{Linguistic Inspired Graph Analysis}

\hypersetup{
pdftitle={Linguistic inspired graph analysis},
pdfauthor={Andrew Broekman, Linda Marshall},
pdfkeywords={Functional Language Theory, Linguistics, Graph Theory, Isomorphisms, Metamodelling, Synatic Information, Semantic Information, Pragmatic Information},
}

\begin{document}
\maketitle

\begin{abstract}
This is an unpublished technical report that introduces a mapping between linguistics and mathematical graph theory. Linguistics and graph theory will be broken down to study the overlap of constituent parts.

Isomorphisms allow human cognition to transcribe a potentially unsolvable problem from one domain to a different domain where the problem might be more easily addressed. Current approaches only focus on transcribing structural information from the source to target structure, ignoring semantic and pragmatic information. Functional Language Theory presents five subconstructs for the classification and understanding of languages. By deriving a mapping between the metamodels in linguistics and graph theory it will be shown that currently, no constructs exist in canonical graphs for the representation of semantic and pragmatic information. It is found that further work needs to be done to understand how graphs can be enriched to allow for isomorphisms to capture semantic and pragmatic information. This capturing of additional information could lead to understandings of the source structure and enhanced manipulations and interrogations of the contained relationships. Current mathematical graph structures in their general definition do not allow for the expression of higher information levels of a source.
\end{abstract}


\section{Introduction}
This report examines whether using graph isomorphisms to create a \textit{target structure} from a \textit{source structure} might exclude certain information, such as syntactic, semantic and pragmatic information. To facilitate the identification process this report considers the domain of linguistics, more specifically looking at Functional Linguistic Theory.  An investigation is conducted to determine whether a possible overlap exists between linguistics and graph theory. A corresponding taxonomy and mapping will be derived between the two domains. A mapping allows for the study of overlap by examining which constructs can be mapped. In the derivation of the the associated mappings the use of visual aids and tables will serve as auxiliary proof~\citep{Thomas2006DevelopingThinking}.

Section~\ref{sec:linguistics} provides a brief overview of linguistics used as a basis for the derivation of the mapping later in the text. Section~\ref{sec:graphs} provides the fundamental definitions for the graph structures, which will be investigated for expansion. The mapping between linguistics and graph theory is presented in Section~\ref{sec:linguisticGraphs}. Future work is discussed in Section~\ref{sec:future}, followed by the conclusion in Section \ref{sec:conclusion}.

\section{Linguistics}
\label{sec:linguistics}

Linguistics can be understood to be the study of human language~\citep{Widdowson1996Linguistics}. Chomsky defines a language ``\textit{to be a set (finite or infinite) of sentences, each finite in length and constructed out of a finite set of elements}"~\citep{Chomsky1957SyntacticStructures}. The definition of Chomsky is provided formally in Definition~\ref{defn:chomskyFormal}.

\begin{leftbar}
\begin{defn}[Chomsky Formal Definition]
\label{defn:chomskyFormal}
Given a finite set E of elements. The subset $S \subseteq E$ is called a \textit{sentence}. A language is defined by $L \coloneqq \{ s \in \dot{S} \vert \dot{S} = \underset{\forall i}{\bigcup} S_{i}, S_{i} \subseteq E\}$
\end{defn}
\end{leftbar}

\begin{figure}
    \centering
    \includegraphics[width=6cm]{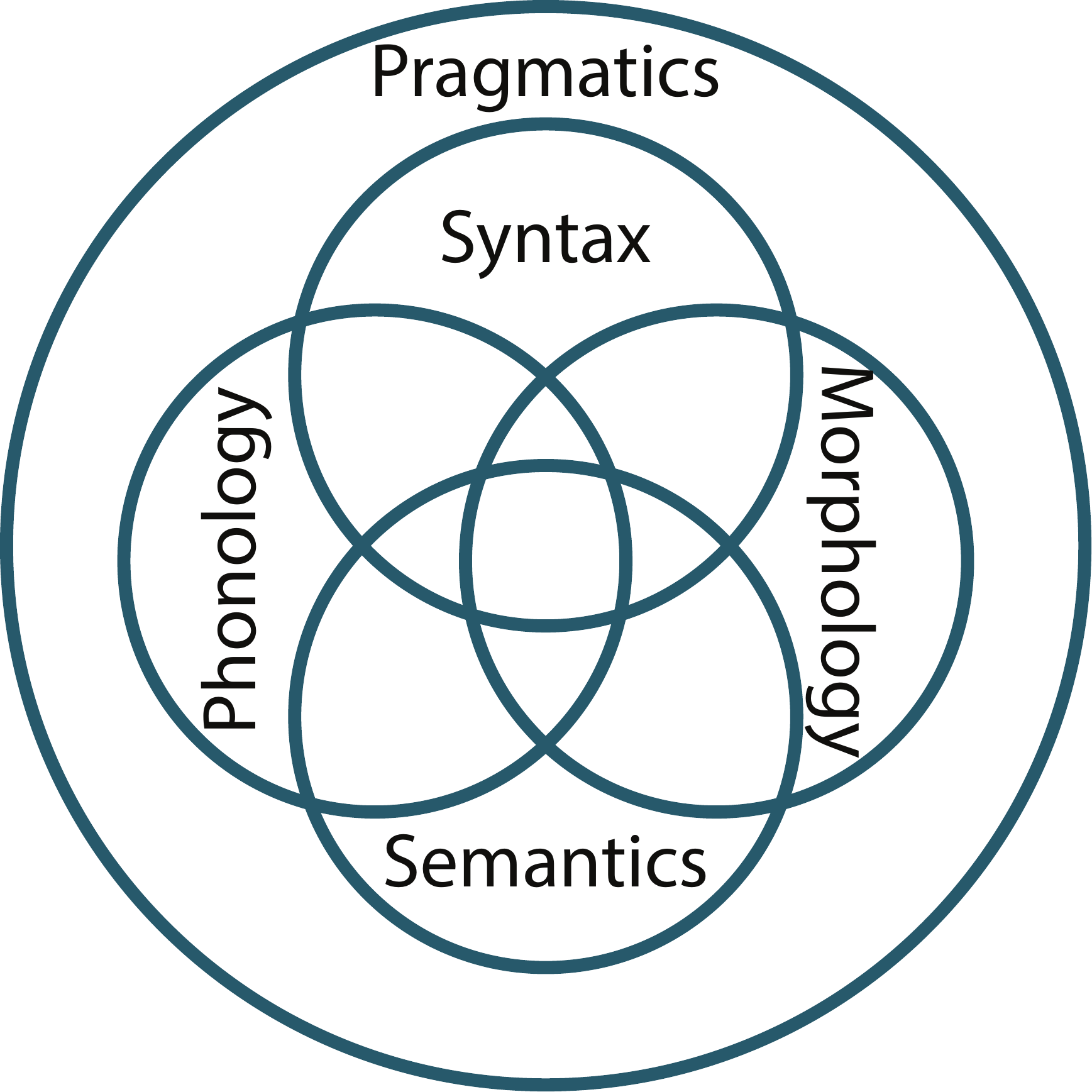}
    \caption{Venn diagram representing functional linguistics by Dr. Md. Enamul Hoque, 2021, retrieved from \url{https://doi.org/10.13140/RG.2.2.28527.07843} Copyright by Education and Development Research Council (EDRC), Bangladesh}
   \label{fig:functionalLinguistics}
\end{figure}

Modern linguistics is based on Functional Language Theory and divides language into five components: (a) \textit{phonology} --- smallest sound unit in a language, i.e. \text{phoneme}; (b) \textit{morphology} --- smallest unit in a language that has meaning, i.e. \textit{morpheme}; (c) \textit{syntax} --- set of rules on how morphemes can be combined into larger expressions; (d) \textit{semantics} --- concerned with the meaning conveyed by syntax; (e) \textit{pragmatics} --- the context surrounding the use of the language~\citep{Widdowson1996Linguistics,Hoque2015ComponentsHoque,HickeyLevelsLanguage}. Figure~\ref{fig:functionalLinguistics} presents a visual representation of Functional Language Theory. This section provides a brief overview of each component and the function that it serves within a language of choice.

During the discussion of linguistics, two meta models are present, namely: (a) \textit{linguistic constructs} --- a mental metamodel of abstract constructs used to model linguistic concepts; (b) \textit{realised linguistic constructs} --- a metamodel of the realisation of various abstract constructs typically by a machine in question.

\subsection{Phonology}
\label{sec:phonology}
The field of phonetics studies the set of all human sounds. Phonetics focuses on the emission, transmission and reception of sound which is termed \textit{articulatory}, \textit{acoustic} and \textit{auditive} phonetics respectively~\citep{Hardcastle2010TheEdition,Hammerstrom2014ArticulatoryDescription,HickeyLevelsLanguage}. Articulatory phonetics studies the sounds brought forth by the sender of information (\textit{the speaker}). The field of auditive phonetics studies how the receiver of information (\textit{the hearer}), receives and interprets the sounds. Acoustic phonetics studies the medium (\textit{the channel}) used to transfer the sounds of the information~\citep{Hammerstrom2014ArticulatoryDescription}. The transference of information using sound needs to occur in a physical medium; examples include sound waves through particles, ink marks on paper or markings on a digital screen. 

In contrast the area of phonology is concerned with a subset of human sounds and the classification of said sounds under a language of study. Phonology furthermore also studies the relationships between phonemes~\citep{Widdowson1996Linguistics,Hoque2015ComponentsHoque,HickeyLevelsLanguage}. Phonemes are defined later in this section. 

A \textit{phoneme} is the smallest abstract unit of sound in a language that has meaning. The phonemes of the word \textit{cool} is given by /k/, /u:/, /l/, where forward slashes are used as delimiters. The phonemes between languages can differ, in other words. a phoneme in one language might not be a phoneme in another language. In English, the \textit{r} is not a phoneme. The reason is that the \textit{r} sound variations, i.e. single, flap, and rolled, do not differentiate the meaning of words but are merely a consequence of the letter's location in a word. In other languages, such as Spanish, the \textit{r} sound can differentiate between words such as [perro] and [pero] representing \textit{dog} and \textit{but} respectively.~\citep{Widdowson1996Linguistics,HickeyLevelsLanguage}

An \textit{allophone} is the phonetic realisation of a phoneme. Various reasons contribute to the different realisations of a phoneme. Reasons include syllable position, surrounding sounds, the distinction between short and long vowels~\citep{Widdowson1996Linguistics,HickeyLevelsLanguage}.

Hickey defines a \textit{phone} as "\textit{the smallest unit of human sound which is recognisable but not classified}". The phones in the word \textit{peat} can be represented as [p], [i:], [t]. Note that square brackets are used as the delimiters for phones.~\citep{Widdowson1996Linguistics,HickeyLevelsLanguage}

\subsection{Morphology}
\label{sec:morphology}
Morphology is the study of words within a language including their internal structure and relationships to one another. Morphology can be subdivided into two fields namely: (a) \textit{word formation} --- changes a word undergoes when altered to form a new word; (b) \textit{grammatical inflection} --- changes a word undergoes when assuming a different role in a sentence~\citep{Widdowson1996Linguistics,HickeyLevelsLanguage}.

A word is some linguistic element characterised by internal stability, and external mobility~\citep{Lyons1968IntroductionLinguistics,HickeyLevelsLanguage}. Internal stability refers to the fact that a word can not be broken down into a set of two or more independent linguistic elements. An example of a word with internal stability is the word \textit{numb}, in that the word can not be broken down into further elements, that is $\textit{n}+\textit{umb}$, $\textit{nu}+\textit{mb}$ or $\textit{num}+\textit{b}$. The word itself is mobile, as it can be used in various syntactic constructions, such as \textit{My arm was numb after the operation} and \textit{Being caught out in the snow numbed my fingers}.

The smallest unit in a language that has meaning is known as a morpheme. Every word is constructed out of one or more morphemes, but morphemes are not necessarily words. The word \textit{cats} is constructed from two morphemes namely: (a) \textit{cat} --- a root morpheme; (b) \textit{-s} --- a morpheme indicating plurality. The morpheme \textit{cat}, in the preceding sentence is a word since it can stand alone. Similar to phonology, the allomorph is the realisation of a morpheme. The morpheme ``ed'' is realised in various phonological allomorphes such as phonentic symbol /\textipa{Id}/ in ``part\^{}ed'', similarly for the /\textipa{d}/ in ``pull\^{}ed'' or /\textipa{t}/ in ``push\^{}ed''.

\subsection{Syntax}
\label{sec:syntax}

\begin{figure}
    \centering
    \includegraphics[width=6cm]{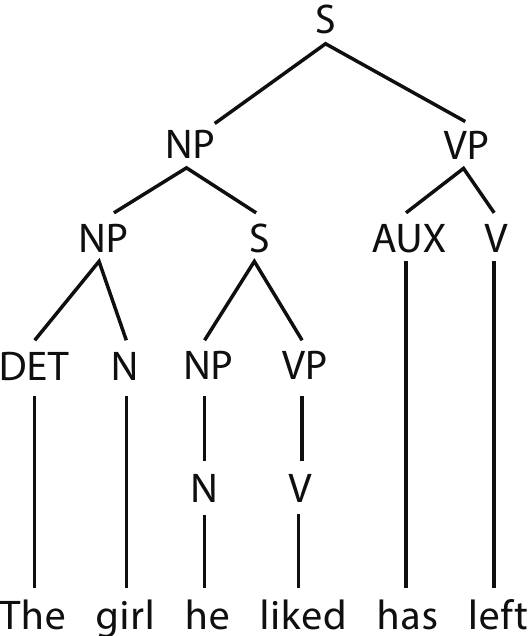}
    \caption{Example of a tree diagram tree representing the underlying deep sentence structure.}
   \label{fig:syntaxTree}
\end{figure}

Syntax is the rule set of the language under study, concerned with how words and morphemes can be combined into larger expression units, often referred to as \textit{sentences}. The only field with comparative research and analysis in linguistics is the field of phonology~\citep{HickeyLevelsLanguage}. \textit{Autonomy of syntax} refers to the rules of the language not needing to be affected by a context outside the language itself, that is to say grammars of languages are autonomous~\citep{Croft1995AutonomyLinguistics,Doetjes1998DegreeSyntax,HickeyLevelsLanguage}.

The analysis of syntax focuses on a few critical areas of sentences. The first focus is on the ordering of the elements within the sentence structure. This ordering is concerned with word classes such as nouns, verbs, pronouns, prepositions and adjectives. The second is to examine sentence structure to explain surface ambiguities and how they arise. Finally is to analyse the relatedness of sentences to one another.

Two structures are studied in syntax analysis: (a) \textit{deep structure} --- refers to the structure not visible in sentences; (b) \textit{surface structure} --- refers to the structure of a sentence in its spoken or written form~\citep{Hockett1958ALinguistics,Chomsky2019DeepInterpretation,HickeyLevelsLanguage}. The syntactic relations between elements of a sentence, such as an object, subject and predicate, are considered deep structures. Deep structures have the characteristic that they can differentiate sentences when no surface structure differences are present. The surface structure of a sentence concerns itself with superficial characteristics, such as the order of elements.

Linguists make use of a variety of tools to study the syntax of sentences. One such tool is to visually represent the underlying structure using a tree diagram, depicted in Figure~\ref{fig:syntaxTree}.
Tree diagrams are often used in the analysis of sentences and can be seen as one possible way of encoding sentences in a particular language. Whether these diagrams accurately reflect mental models is debatable. The use of tree diagrams highlights the abstract properties of: (a) \textit{temporal  precedence} --- concerned with which elements precede other elements; (b) \textit{dominance relations} --- concerned with the relations that various units have with regards to one another~\citep{Cornell2014ModelSyntax,HickeyLevelsLanguage}.

\subsection{Semantics}
\label{sec:semantics}
The morphemes, the structural unit of a language, are concerned with expressing an idea and serve a function in the language. The formal definition of a language, Definition \ref{defn:chomskyFormal}, ignores this fact \citep{Widdowson1996Linguistics}. This omission raises the need to know more about the words, not as structural concepts, rather as concepts that convey meaning together with syntax referred to as semantics. Semantics is concerned with the study of the encoded meaning in a sentence and can be thought of as a \textit{functional grammar} \citep{Widdowson1996Linguistics,Feng2013FunctionalLearning,HickeyLevelsLanguage}. 



\subsection{Pragmatics}
\label{sec:pragmatics}
Given a piece of text, say $T$, various questions can be asked about the text $T$. Who wrote the text? Why was the text written? What is the information conveyed by the text $T$? Who is the intended receiver of the information? Thus, pragmatics studies the meaning of information conveyed by a sender and receiver under a specified context~\citep{Widdowson1996Linguistics,HickeyLevelsLanguage}. Examples of context include day-to-day context, emergency context, workplace context, to name a few. 


\section{Graphs}
\label{sec:graphs}

\begin{figure}
     \centering
     \begin{subfigure}[b]{0.3\textwidth}
         \centering
         \includegraphics[width=\textwidth]{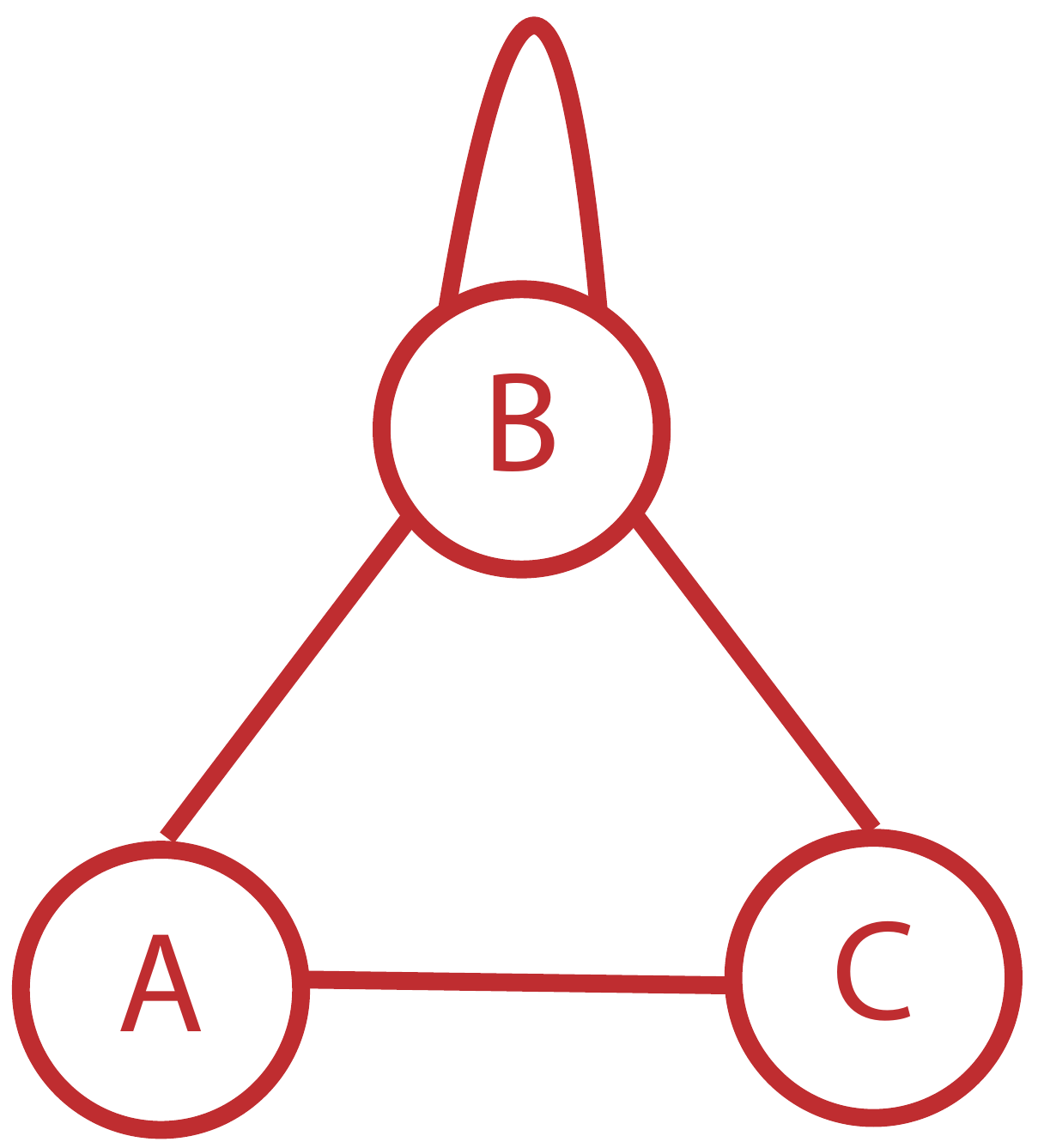}
         \caption{Mental representations of an undirected graph}
         \label{subfig:undirectedGraph}
     \end{subfigure}
     \begin{subfigure}[b]{0.3\textwidth}
         \centering
         \includegraphics[width=\textwidth]{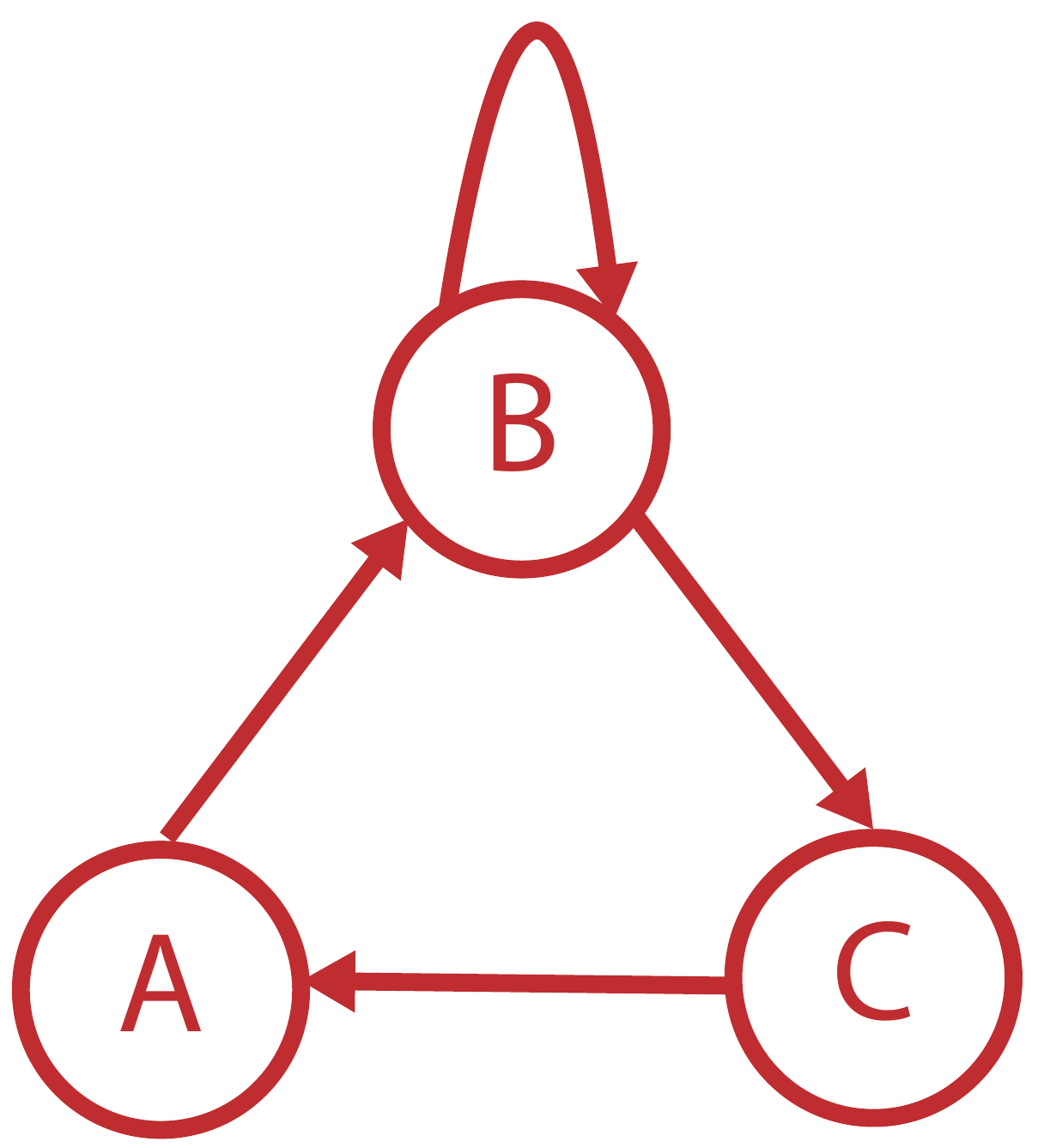}
         \caption{Mental representations of a directed graph}
         \label{subfig:directedGraph}
     \end{subfigure}
        \caption{Examples of mental metamodels of mathematical graph structures}
        \label{fig:graphMentalModels}
\end{figure}

Graph structures are mathematical structures used to represent relationships between two or more entities. These structures provide various disciplines with a concrete mechanism to express, manipulate, and interrogate relationships. Graphs have found a place among social network analysis \citep{Wasserman1994SocialApplications}, web page indexing \citep{Page1999TheWeb.}, metabolic analysis \citep{Jeong2000TheNetworks} and knowledge representation \citep{Sowa2014PrinciplesKnowledge}. More recent applications, include the epidemiological study and visualisation of infectious disease, especially related to the \acrfull{covid19}~\citep{Verspoor2020COVID-SEE:Research,Bras2020VisualisingResearch,Cernile2021NetworkDiscovery}.

A graph structure consists of two sets: a set of vertices $V$ and a set of edges $E$. The canonical form is presented first, also known as an undirected graph or simply a graph. Following this, the concept of directional edges is introduced, which defines the directed graph or digraph. This report presents a brief overview of graph structures. The interested reader is directed towards \textit{Graph Theory with Applications} by Bondy and Murty (1976) and \textit{Graph Theory} by Diestel (2017) for a more in-depth discussion on graph theory.

The representations of the graphs given in Figure~\ref{fig:graphMentalModels} are best understood to be mental metamodel constructions as represented by a machine\footnote{In the provided context machine refers to an entity that can compute. Therefore either a human or computer can be considered.}. In contrast to the realisations by some machines of these same graphs provided in Figure~\ref{fig:graphRealisations}.

\subsection{Undirected Graph}
Various authors provide different definitions for a undirected graph structure. The first definition of Bondy \& Murty makes use of an incidence function to define the edges completely. The definition of Bondy \& Murty is reproduced in Definition~\ref{defn:graphBondyMurty} \citep{Bondy1976GraphApplications}.

\begin{leftbar}
\begin{defn}[Undirected Graph - Bondy \& Murty, 1976]
\label{defn:graphBondyMurty}
A graph G is defined as an ordered triple $(V(G),E(G), \phi_G)$ where:
\begin{itemize}
    \item V(G) is a non-empty set of vertices
    \item E(G) is a set disjoint from V(G) of edges, and
    \item $\phi_G$ is an \textit{incidence function} that associates with each edge of G an unordered pair of vertices G, i.e. $\phi_G : E \to \{U \subseteq V \mid 1 \leq \vert U \vert \leq 2\}$
\end{itemize}
\end{defn}
\end{leftbar}

Specification~\ref{spec:undirectedGraph} is obtained by applying Definition~\ref{defn:graphBondyMurty} on Figure~\ref{subfig:undirectedGraph}.

\begin{leftbar}
\begin{spec}
\begin{equation*}
\label{spec:undirectedGraph}
\begin{split}
V(G) & = \{A, B, C \},\\
E(G) & = \{e_1, e_2, e_3, e_4 \}\\
\phi(e_1) & = \{ A, B \},\\
\phi(e_2) & = \{ B, C \},\\
\phi(e_3) & = \{ C, A \},\\
\phi(e_4) & = \{ B, B \}\\
          & = \{ B \}
\end{split}
\end{equation*}
\end{spec}
\end{leftbar}





\subsection{Directed Graph}
When directionality is introduced onto the edges of a graph G, a directed graph or digraph structure emerges. For the directed graph G, direction on the edges implies that the edge from vertex $a$ to vertex $b$ is different to the edge going from vertex $b$ to vertex $a$ for $\forall a, \forall b \in V(G)$. To allow for directionality in graphs the incidence function can be redefined to map an edge to a set of ordered pairs. The incidence function in Defintion~\ref{defn:graphBondyMurty} can be modified to a define a directed graph, as per Definition~\ref{defn:digraphBondyMurty}.

\begin{leftbar}
\begin{defn}[Directed Graph - Bondy \& Murty, 1976]
\label{defn:digraphBondyMurty}
A digraph G is defined as an ordered triple $(V(G),E(G), \phi_G)$ where:
\begin{itemize}
    \item V(G) is a non-empty set of vertices
    \item E(G) is a set disjoint from V(G) of edges, and
    \item $\phi_G$ is an \textit{incidence function} that associates with each edge of G an ordered pair of vertices G, i.e. $\phi_G : E \to \{U \subseteq V \times V \}$
\end{itemize}
\end{defn}
\end{leftbar}

Specification~\ref{spec:directedGraph} is obtained by using Definition~\ref{defn:digraphBondyMurty} on Figure~\ref{subfig:directedGraph}.

\begin{leftbar}
\begin{spec}
\begin{equation*}
\label{spec:directedGraph}
\begin{split}
V(G) & = \{A, B, C \},\\
E(G) & = \{e_1, e_2, e_3, e_4 \},\\
\phi(e_1) & = ( A, B ),\\
\phi(e_2) & = ( B, C ),\\
\phi(e_3) & = ( C, A ),\\
\phi(e_4) & = ( B, B )\\
\end{split}
\end{equation*}
\end{spec}
\end{leftbar}

\section{Linguistic Graphs}
\label{sec:linguisticGraphs}

\begin{figure}
     \centering
     \includegraphics[width=12cm]{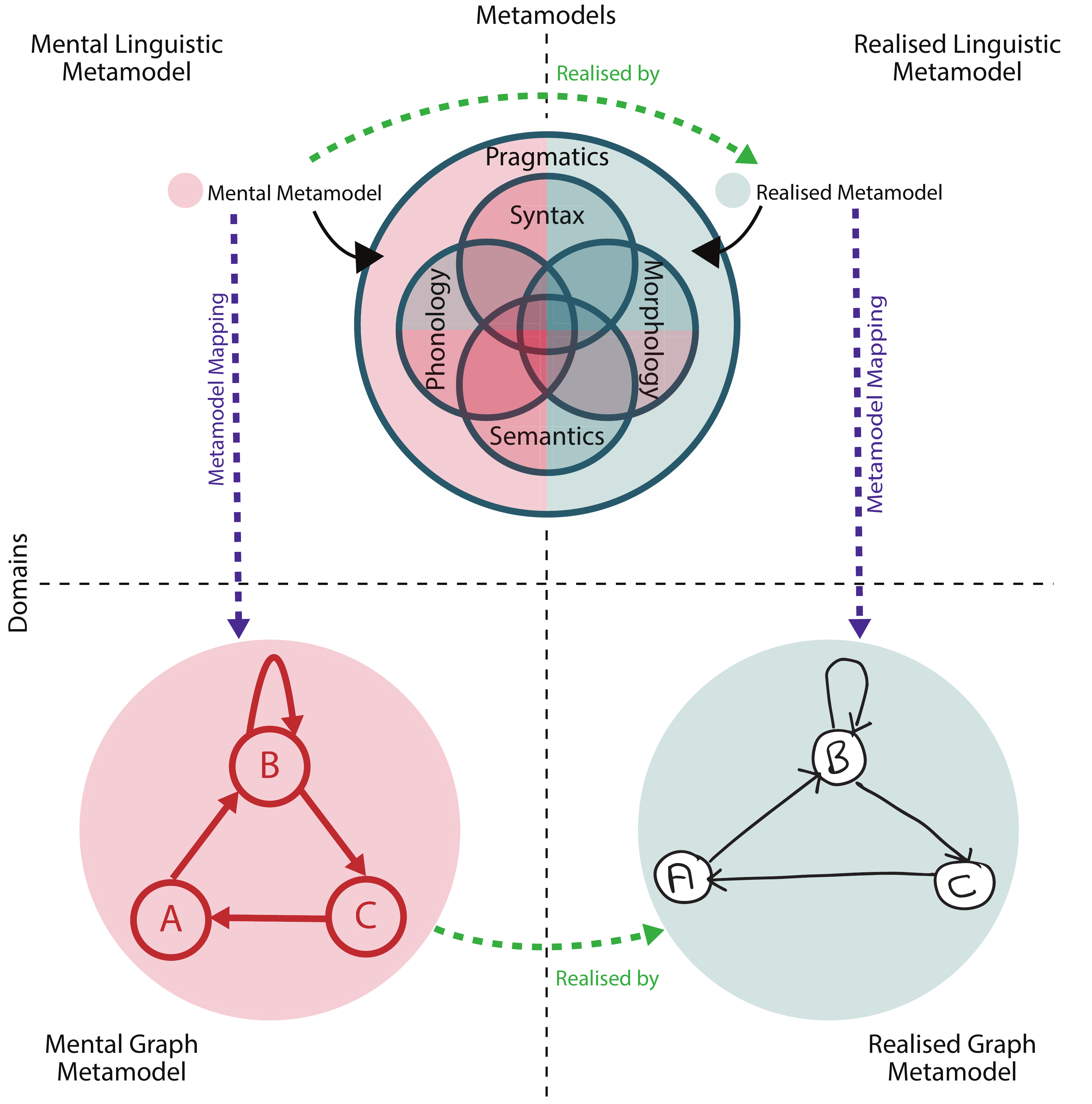}
     \caption{Visual representation of the domains, metamodels and mappings considered in this report}
     \label{fig:mappingGraph}
\end{figure}

Section \ref{sec:linguistics} presented the reader with various concepts around linguistics while Section~\ref{sec:graphs} presented a brief overview of graph theory. This section investigates the relationships between various linguistics components and graphs and derives associated mappings between the two domains. The visual reader can refer to Figure~\ref{fig:mappingGraph} for a more detailed analysis of the purpose of this report. 

Figure~\ref{fig:mappingGraph} illustrates two domains: (a) \textit{linguistic} - the domain concepts located at the top of the figure; (b) \textit{mathematical graphs} - the domain concepts located at the bottom of the figure. Within each of the listed domains, two metamodels exist: (a) \textit{mental metamodel} ---  what a machine understands abstractly about the domain, located on the left hand side of the figure; (b) \textit{realised metamodel} --- metamodel of the realised mental metamodel located on the right hand side of the figure. The observant reader will notice that the transmission of information in the linguistic domain is normally done via sound in contrast to the visual transmission utilised by graphs.

The functional linguistics representation from Figure \ref{fig:functionalLinguistics} has been coloured to highlight that each of the subconstructs of Fundamental Language Theory is composed out of two metamodels. the mental and realised metamodel indicated by red and teal transparent half-circles respectively. The mathematical graph domain is presented as two individual red and teal circles, one for each metamodel. The realisation mappings, represented by the green dashed line, exist implicitly in both domains. 

This report proceeds to investigate and derive mappings between the metamodel represented by the dashed purple line. One mapping between the mental metamodel of the five language subcomponents and the mental metamodel of graphs. The second mapping is between the realised metamodel of the five language subcomponents and the realised metamodel of graphs. In deriving the mapping, the macroscopic characteristics of linguistics and graphs are first considered and mapped at the syntactic level. Thereafter the mapping is refined by moving to more abstract concepts in linguistics, that is moving from the syntax subconstruct towards the phonology subconstruct.

\subsection{Syntax}

\begin{table}
    \centering
    \begin{tabularx}{\textwidth}[t]{XX}
        \hline \hline
        & \textbf{Syntax} \\
        \hline \hline
        
        \begin{minipage}[t]{\linewidth}%
        Syntax\\
        \textit{``The constituent structure of sentences''} \citep{Widdowson1996Linguistics}
        \end{minipage}
        & 
        \begin{minipage}[t]{\linewidth}%
        Incidence function\\
        \textit{Incidence function provides the rules as to how edges may connect, i.e. only single or two vertex in normal graph, or multiple vertices in hypergraphs}
        \end{minipage}\\ \hline
        
        \begin{minipage}[t]{\linewidth}%
        Ordering of elements\\
        \textit{General description of a phone}
        \end{minipage}
        & 
        \begin{minipage}[t]{\linewidth}%
        Ordered graph\\
        \textit{A graph structure with a total or partial ordering}
        \end{minipage}\\ \hline
        
        \begin{minipage}[t]{\linewidth}%
        Empty Categories\\
        \textit{A linguistic category assumed to exist in a sentence but without any manifestation} \citep{HickeyLevelsLanguage}.
        \end{minipage}
        & 
        \begin{minipage}[t]{\linewidth}%
        Ordered graph\\
        \textit{A graph structure with a total or partial ordering}
        \end{minipage}\\ \hline
        
        \begin{minipage}[t]{\linewidth}%
        Surface ambiguities\\
        \textit{Ambiguity arising from the speaker}\citep{HickeyLevelsLanguage}
        \end{minipage}
        & 
        \begin{minipage}[t]{\linewidth}%
        ---
        \end{minipage}\\ \hline
        
        \begin{minipage}[t]{\linewidth}%
        Relatedness of sentences\\
        \textit{Indicates the relatedness of sentences by using movement rules as notational means} \citep{HickeyLevelsLanguage}.
        \end{minipage}
        & 
        \begin{minipage}[t]{\linewidth}%
        Graph difference\\
        \textit{Graph difference as proposed by Marshall (2014)}
        \end{minipage}\\ \hline
        
        \begin{minipage}[t]{\linewidth}%
        Surface structure\\
        \textit{Actual form of a spoken or written sentence} \citep{HickeyLevelsLanguage}.
        \end{minipage}
        & 
        \begin{minipage}[t]{\linewidth}%
        Graph G\\
        \textit{A graph given by G, the visual or mental realisation of the graph}
        \end{minipage}\\ \hline
        
        \begin{minipage}[t]{\linewidth}%
        Deep structure\\
        \textit{Level of representation where the meaning of the sentence structure is unambiguous} \citep{HickeyLevelsLanguage}.
        \end{minipage}
        & 
        \begin{minipage}[t]{\linewidth}%
        ---
        \end{minipage}\\ \hline
        
        \begin{minipage}[t]{\linewidth}%
        Temporal precedence\\
        \textit{Reflects the precedence relations of the elements in the sentence} \citep{HickeyLevelsLanguage}.
        \end{minipage}
        & 
        \begin{minipage}[t]{\linewidth}%
        Total or partial ordering (Horizontal) \\
        \textit{A total or partial ordering define on the graph G}
        \end{minipage}\\ \hline
        
        \begin{minipage}[t]{\linewidth}%
        Dominance relations\\
        \textit{Relation of elements to each other in sentences} \citep{HickeyLevelsLanguage}.
        \end{minipage}
        & 
        \begin{minipage}[t]{\linewidth}%
        Total or partial ordering (Vertical) \\
        \textit{A total or partial ordering define on the graph G}
        \end{minipage}\\ \hline
        
        \begin{minipage}[t]{\linewidth}%
        Encoding\\
        \textit{Various forms of encoding sentences are possible, e.g. tree diagrams} \citep{HickeyLevelsLanguage}.
        \end{minipage}
        & 
        \begin{minipage}[t]{\linewidth}%
        Encoding (Computational representation) \\
        \textit{The method in which the graph is encoded, e.g. adjacency list, adjacency matrix, incidence matrix  or object oriented representation \citep{Drozdek2013DataJava}}
        \end{minipage}\\ \hline
    \end{tabularx}
    \caption{Mapping from linguistic syntax concepts to corresponding concepts in graphs}
    \label{tbl:mappingSyntax}
\end{table}

The mapping of syntactic concepts to graphs is studied, after which the mapping of morphology and phonology will be considered. Section \ref{sec:linguistics} identifies that linguistic syntax has the sentence as a core construct which is composed out of smaller units, namely morphemes. From Definition \ref{defn:graphBondyMurty}, a graph structure is comprised out of two smaller units, namely vertices and edges. The first natural mapping is that a node and a vertex is to graph theory what the morpheme is to linguistics. A graph's construction comprises two building blocks, similar to how sentences are constructed from morphological building blocks. The second mapping arises from observing that the graph fulfils a similar function as the concept of a sentence in linguistic syntax. A graph and sentence is a larger unit of expressing meaning.

The incidence function, as defined by Definition~\ref{defn:graphBondyMurty}, associates with each edge in the graph a set or ordered pair of vertices, depending on whether the graph is directed or not. The incidence function is represented by $\phi_G : E \to \{U \subseteq V \mid 1 \leq \vert U \vert \leq 2\}$ or $\phi_G : E \to \{U \subseteq V \times V \}$ for undirected and directed graphs respectively. The incidence function define stringent constraints on how the units of a graph can be arranged. Explicitly an edge is allowed only to connect vertices, this means  an edge cannot connect to another edge. Hence the incidence function fulfills a similar role as linguistic syntax, defining the rules for a language.

If one considers a graph in terms of set theory, it is natural to define some total or partial ordering of the elements. When an ordering is defined over a graph G, a ordered graph is obtained. This is similar to when syntax defines the ordering of word types onto a language. The ordering of elements map naturally onto the concept of a order relation. Furthermore from set theory it follows that an infimum and supremum can be obtained for the set of vertices and edges. In Section~\ref{sec:syntax}, the temporal precedence and dominance relations hold very close ties to the ordering of linguistic elements. Similarly consider a tree structure, as a specialised graph structure. In a vertical arrangement, where a node has a single parent and child nodes, it is natural to define dominance relations vertically. Similarly, the relation between sibling nodes map naturally onto temporal precedence of a language. The physical or mental model of a graph G can be naturally mapped onto the surface structure of a sentence. To determine the difference in graphs a difference operator can be defined. The difference in graphs would also show the relatedness of graphs, similar to syntax. 

\subsection{Morphology}

\begin{table}
    \centering
        \begin{tabularx}{\textwidth}[t]{XX}
        \hline \hline
        & \textbf{Morphology} \\
        \hline \hline
        
        \begin{minipage}[t]{\linewidth}%
        Morpheme\\
        \textit{``An abstract element of meaning''}\citep{Widdowson1996Linguistics}.
        \end{minipage}
        & 
        \begin{minipage}[t]{\linewidth}%
        Concept of vertex or edge\\
        \textit{Mental metamodel representing the concept of a vertex as a unit of containment and edge as unit of connection between concepts}.
        \end{minipage}\\ \hline
        
        \begin{minipage}[t]{\linewidth}%
        Allomorph\\
        \textit{``The Version of a morpheme as actually realised in speech or writing''} \citep{Widdowson1996Linguistics} 
        \end{minipage}
        & 
        \begin{minipage}[t]{\linewidth}%
        Vertex or Edge\\
        \textit{Realisation of some marking on a medium representing a vertex or edge}
        \end{minipage}\\ \hline
        
        \begin{minipage}[t]{\linewidth}%
        Internal stability\\
        \textit{Word which can not be broken down into further elements} \citep{HickeyLevelsLanguage}.
        \end{minipage}
        & 
        \begin{minipage}[t]{\linewidth}%
        Atomic vertex or edge\\
        \textit{Vertex or edge is unable to be subdivide into smaller representations}.
        \end{minipage}\\ \hline
        
        \begin{minipage}[t]{\linewidth}%
        External mobility\\
        \textit{Words can occupy various positions in a sentence} \citep{HickeyLevelsLanguage}.
        \end{minipage}
        & 
        \begin{minipage}[t]{\linewidth}%
        Vertex movement\\
        \textit{Vertex can be moved around the structure as long as incidence function is obey}
        \end{minipage}\\ \hline
    \end{tabularx}
    \caption{Mapping from linguistic morphology concepts to corresponding concepts in graphs}
    \label{tbl:mappingMorphology}
\end{table}

A vertex and an edge are atomic units of a graph structure, exhibiting internal stability similar to linguistics. Internal stability of a vertex and edge refers to the concept that these units can not be subdivided into smaller units.

External mobility refers to the ability of a word to be placed in various positions in a sentence. The moving of a word does change the structure of a sentence; a vertex and an edge do exhibit external mobility. The induction of a structural change by the movement of a vertex or edge in a graph is consistent with linguistic behaviour.

Recall that the allomorph is the realisation of the morpheme. A vertex or edge realised by markings on a medium is similar to the allomorph for morphemes. The morpheme maps naturally to the mental metamodel a machine has about the abstract concept of containment and the connection between two concepts.

\subsection{Phonology}

\begin{table}
    \centering
    \begin{tabularx}{\textwidth}[t]{XX}
        \hline \hline
        & \textbf{Phonology} \\
        \hline \hline
        
        \begin{minipage}[t]{\linewidth}%
        Phonetic\\
        \textit{``Concerning the actual pronunciation of speech sounds''} \citep{Widdowson1996Linguistics}.
        \end{minipage}
        & 
        \begin{minipage}[t]{\linewidth}%
        Set element\\
        \textit{Mental metamodel of an abstract concept or entity as represented by set theory}
        \end{minipage}\\ \hline
        
        \begin{minipage}[t]{\linewidth}%
        Phoneme\\
        \textit{``The abstract element of sound, identified as being distinctive in a particular language''} \citep{Widdowson1996Linguistics}.
        \end{minipage}
        & 
        \begin{minipage}[t]{\linewidth}%
        Abstract marking\\
        \textit{Abstract marking concept serving to separate the meaning between an entity and some relationship}
        \end{minipage}\\ \hline
        
        \begin{minipage}[t]{\linewidth}%
        Phone\\
        \textit{``Smallest unit of human sound which is recognisable but not classified''} \citep{HickeyLevelsLanguage}.
        \end{minipage}
        & 
        \begin{minipage}[t]{\linewidth}%
        Realised marking\\
        \textit{Realised expression of an entity or relationship}.
        \end{minipage}\\ \hline
        
        \begin{minipage}[t]{\linewidth}%
        Allophone\\
        \textit{``The version of the phoneme as actually realized phonetically in speech''} \citep{Widdowson1996Linguistics}.
        \end{minipage}
        & 
        \begin{minipage}[t]{\linewidth}%
        Difference in marking\\
        \textit{Realised difference in marking to express the concept of an entity or relationship}.
        \end{minipage}\\ \hline \hline
    \end{tabularx}
    \caption{Mapping from linguistic phonology concepts to corresponding concepts in graphs}
    \label{tbl:mappingPhonology}
\end{table}

Linguistic phonology is concerned with sound, however graph structures are conveyed visually. In this section an additional metamapping between sound and visual communication is considered. 

Section \ref{sec:phonology} introduced the concepts of \textit{articulatory}, \textit{acoustic} and \textit{auditive} phonetics. The Shannon-Weaver model of communication presents the idea of a sender, receiver and channel of communication \citep{ShannonWeaver1949}. A metamapping between audio and visual communication can be defined as follows: (a) the machine producing visual markings to transfer information, maps to the concept of articulatory phonetics; (b) the medium on which the visual markings are made, corresponds to acoustic phonetics; (c) the machine processing the visual markings with the goal to receive information, maps to acoustic phonetics.

\begin{figure}
     \centering
     \begin{subfigure}[b]{0.3\textwidth}
         \centering
         \includegraphics[width=\textwidth]{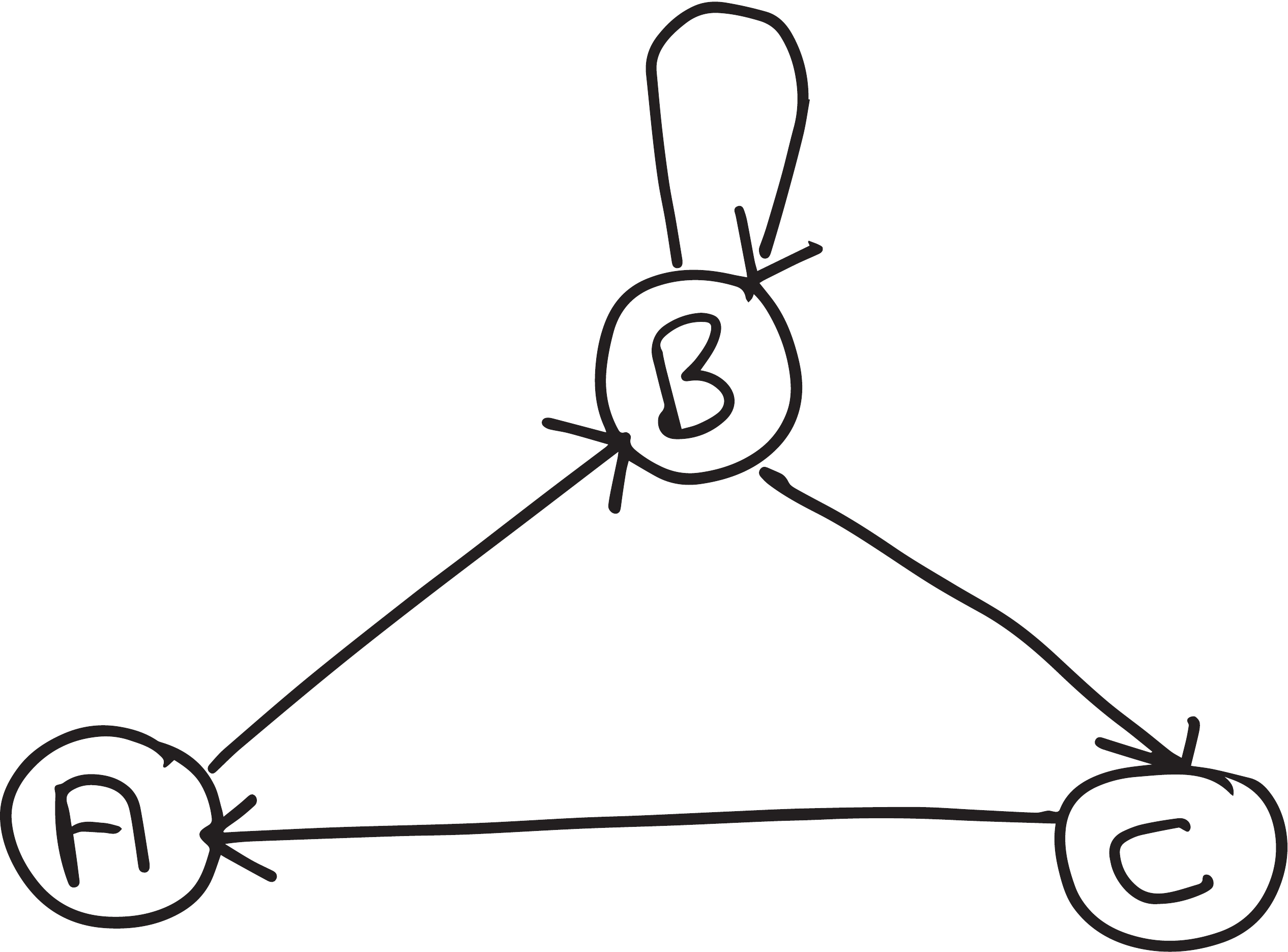}
         \caption{Right handed individual on Wacom tablet input}
         \label{subfig:electronicDrawing}
     \end{subfigure}
     \hfill
     \begin{subfigure}[b]{0.3\textwidth}
         \centering
         \includegraphics[width=\textwidth]{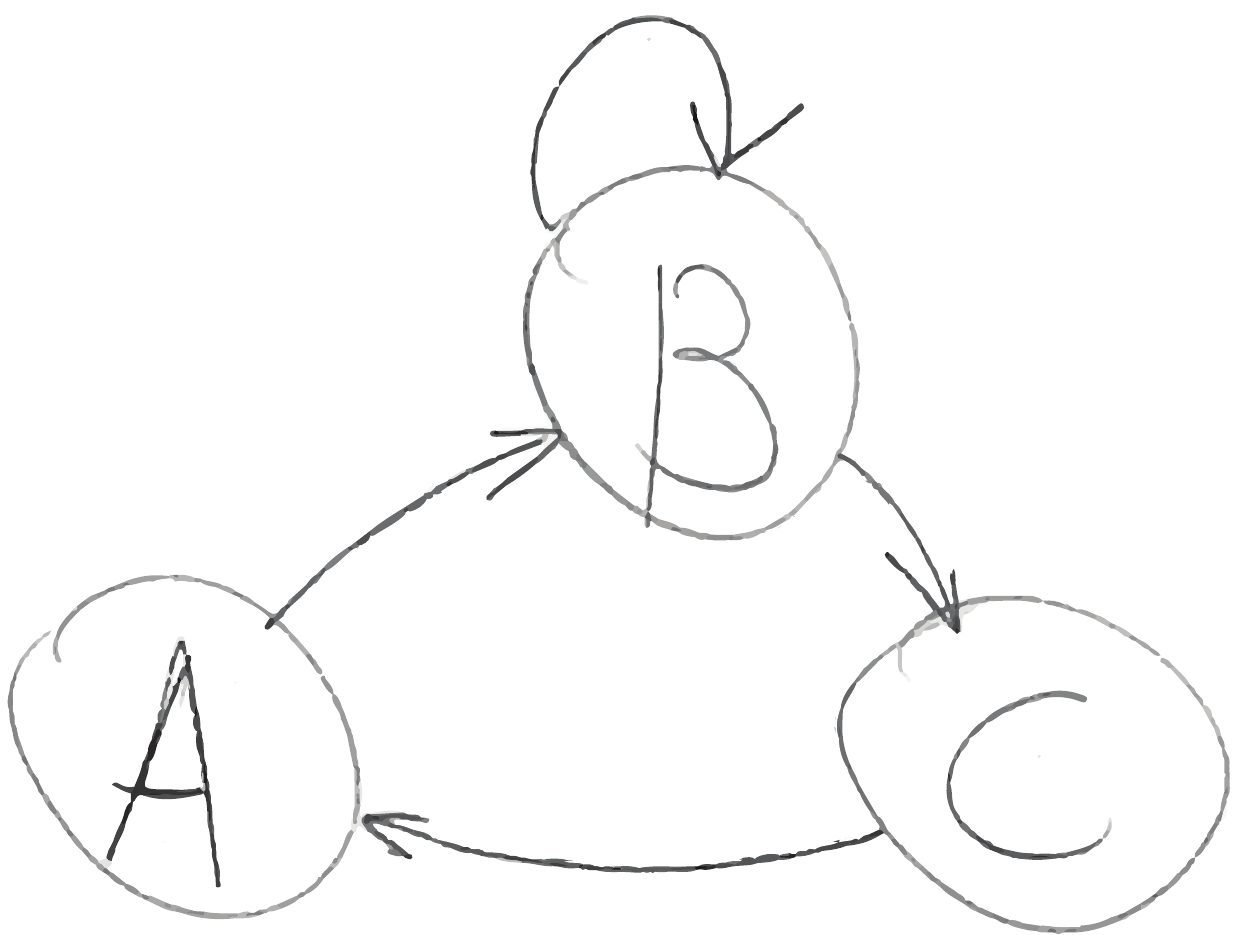}
         \caption{Left handed individual with ballpoint pen on paper}
         \label{subfig:leftHandedBallpointPaper}
     \end{subfigure}
     \hfill
     \begin{subfigure}[b]{0.3\textwidth}
         \centering
         \includegraphics[width=\textwidth]{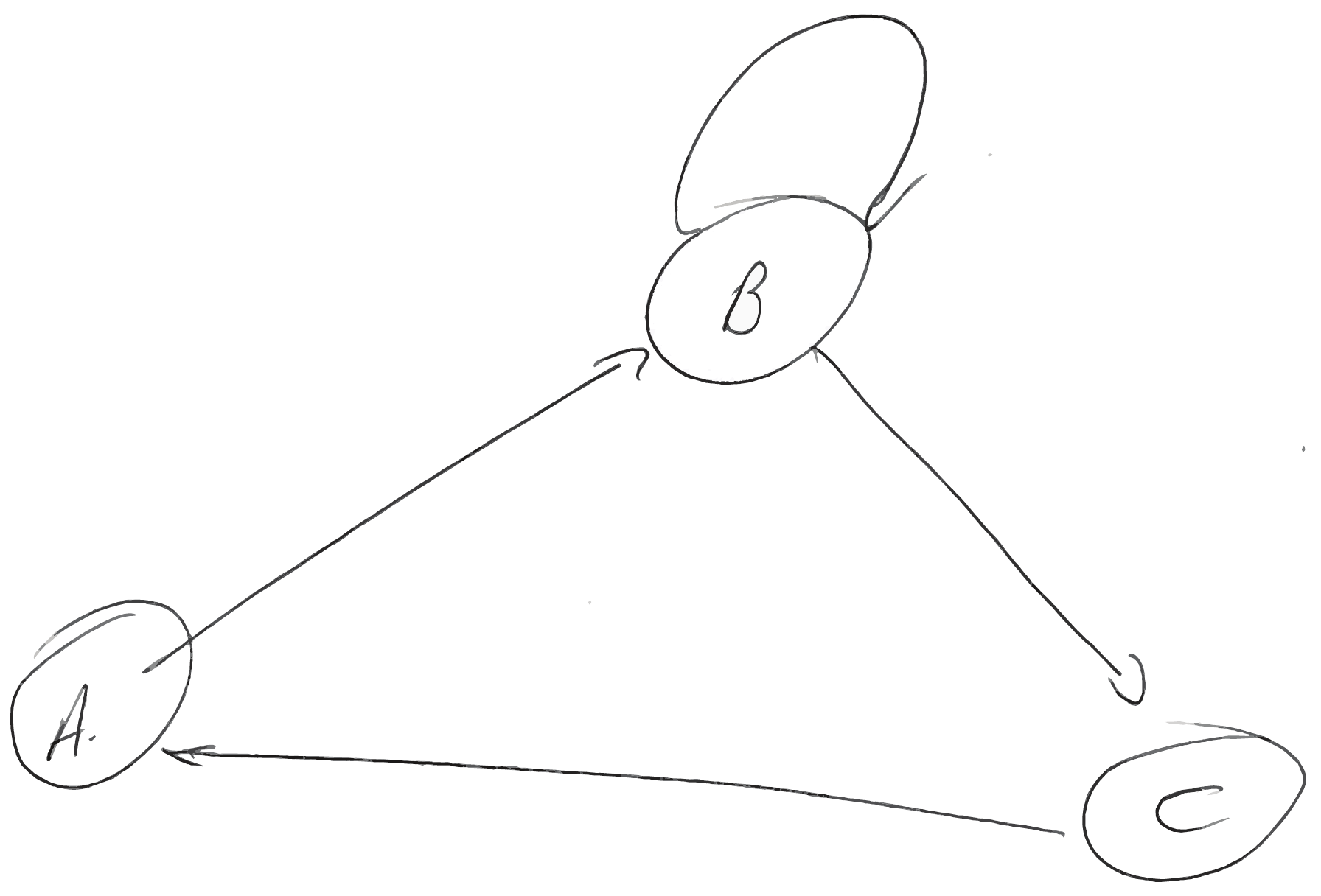}
         \caption{Right handed individual with ballpoint pen on paper}
         \label{subfig:subfig:rightHandedBallpointPaper}
     \end{subfigure}
        \caption{Different realisations of the same abstract graph}
        \label{fig:graphRealisations}
\end{figure}

A phoneme refers to the mental metamodel of a sound representation. The visual equivalent would be the mental metamodel of markings on some medium, that is the mental metamodel of shapes and lines. A phone, the realised sound a human produces, maps naturally onto the realised drawings produced by a human. The realised markings can be any visual marking on a medium, including drawings on a digital screen. The allophone is the different ways that a phoneme can be reproduced. Humans produce different markings due to various aspects, including pressure, stroke types, medium characteristics and mental state, to name a few. The difference in these markings maps onto the allophone. Figure~\ref{subfig:directedGraph} presents a mental model of a graph which is realised by the respective markings in  Figure~\ref{fig:graphRealisations}. The realisations of the graph was made by three different machines, using a variety of mediums.

\subsection{Review}
\begin{figure}
     \centering
     \begin{subfigure}[b]{\textwidth}
         \centering
          \includegraphics[width=\textwidth]{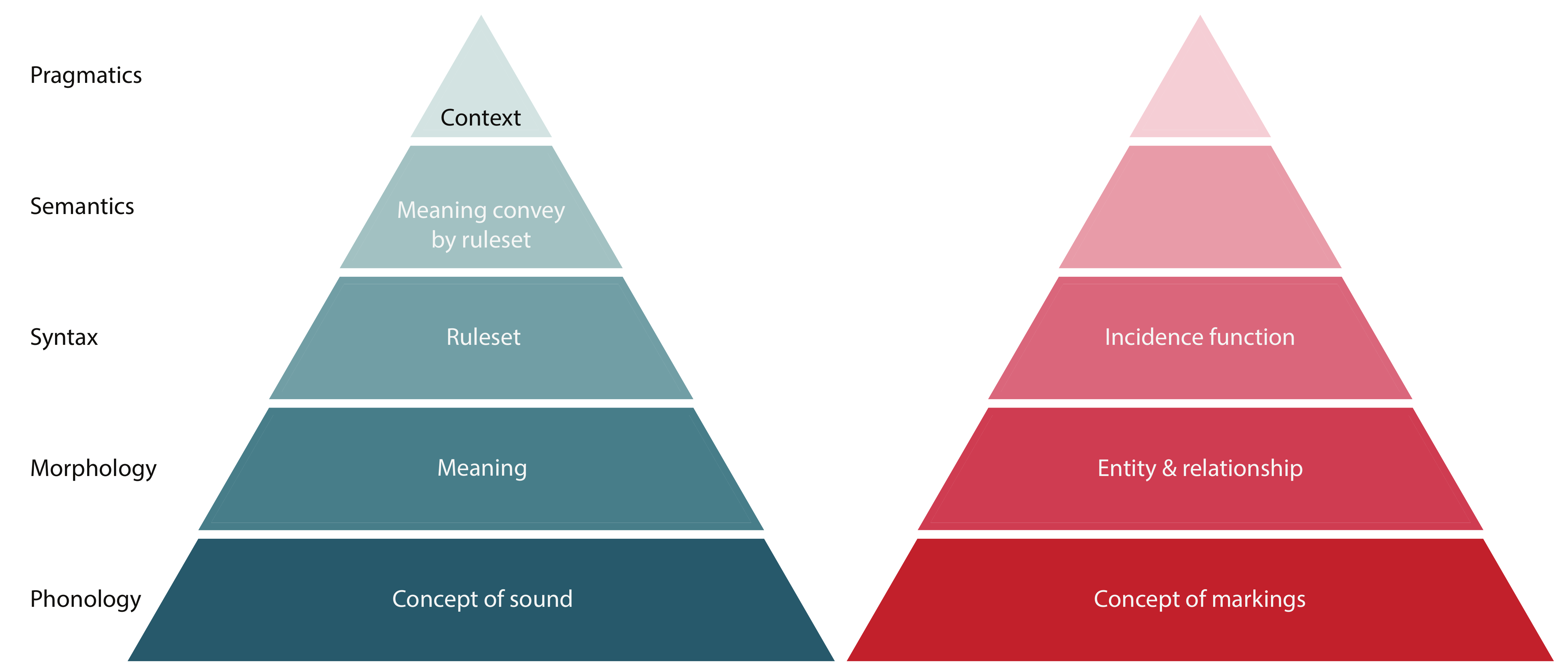}
         \caption{Linguistic and graph \textit{mental} metamodel mappings}
         \label{subfig:mentalMapping}
     \end{subfigure}
     \hfill
     \begin{subfigure}[b]{\textwidth}
         \centering
         \includegraphics[width=\textwidth]{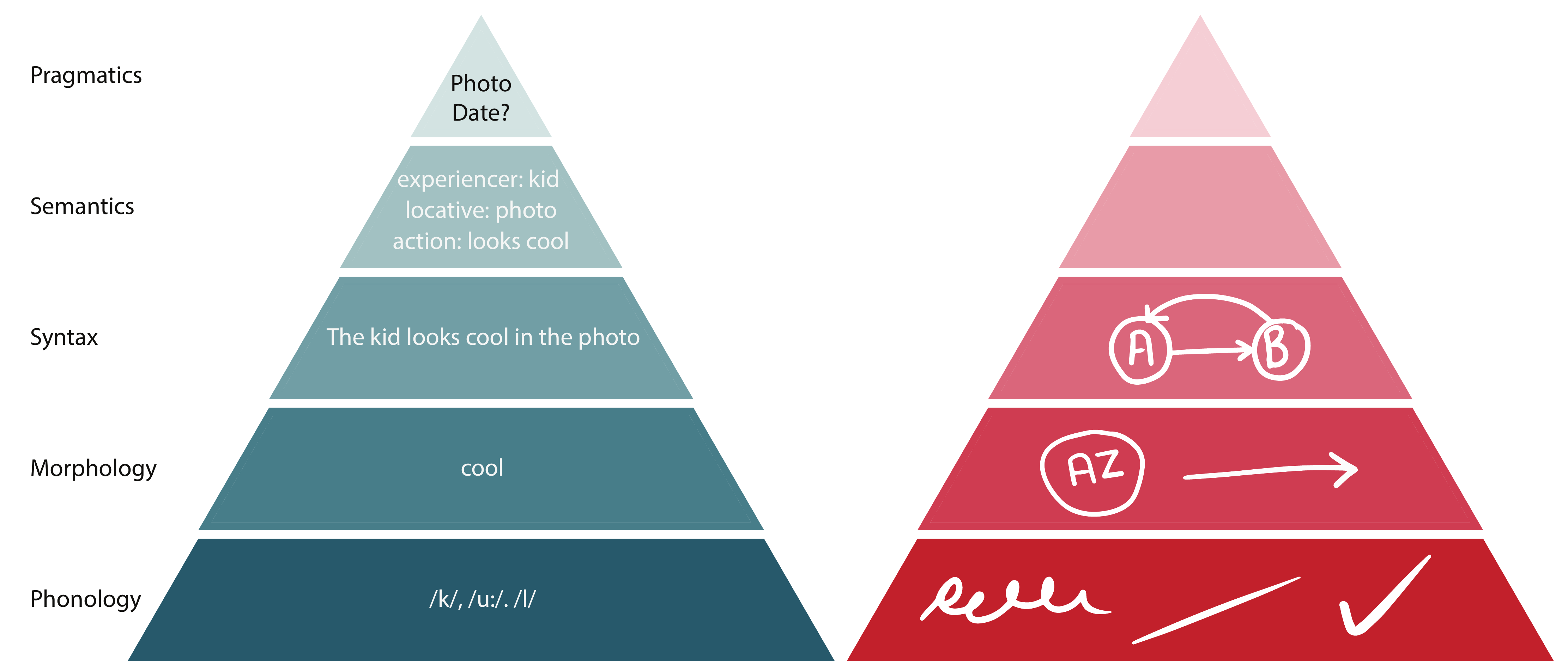}
         \caption{Linguistic and graph \textit{realised} metamodel mappings}
         \label{subfig:realisedMapping}
     \end{subfigure}
     \caption{Taxonomies showing the corresponding mappings between the various metamodels for the machine}
     \label{fig:compareTax}
\end{figure}

\begin{figure}
     \centering
     \begin{subfigure}[b]{0.4\textwidth}
         \centering
         \includegraphics[width=\textwidth]{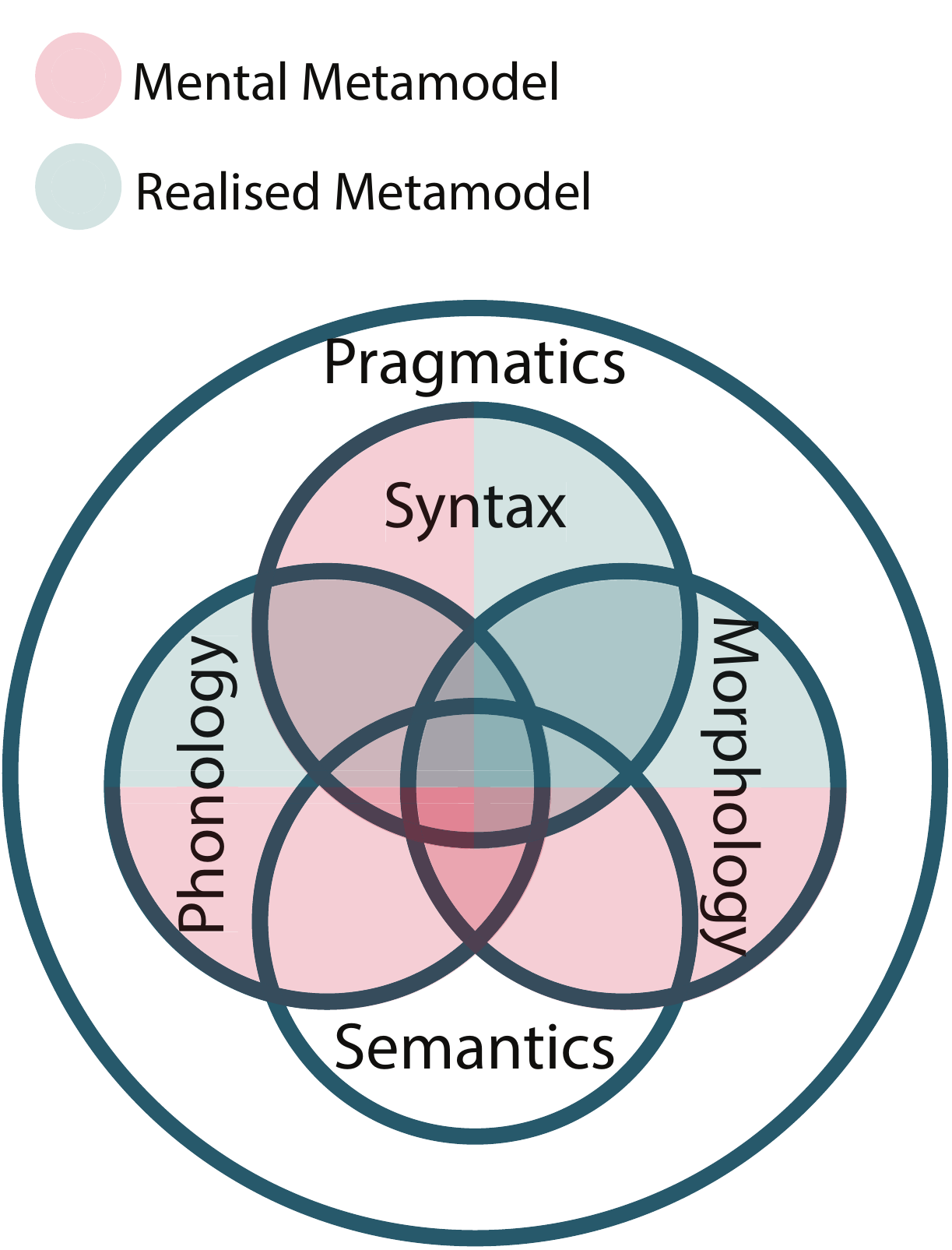}
         \caption{Linguistic component metamodels with \textit{corresponding} mappings to graphs}
         \label{subfig:presentMapping}
     \end{subfigure}
     \hfill
     \begin{subfigure}[b]{0.4\textwidth}
         \centering
         \includegraphics[width=\textwidth]{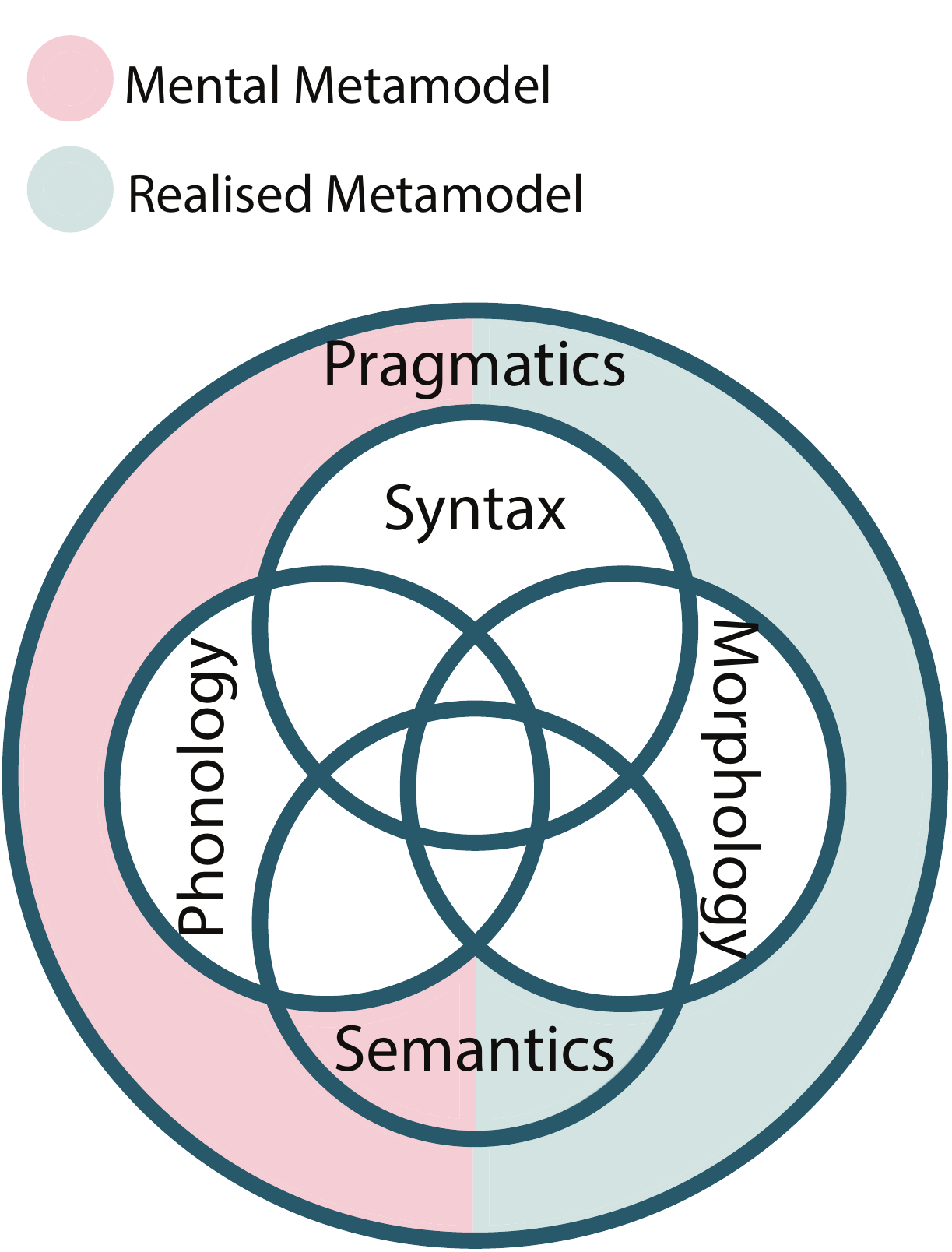}
         \caption{Linguistic component metamodels with \textit{no corresponding} mappings to graphs}
         \label{subfig:missingMapping}
     \end{subfigure}
     \caption{Visual representation of the metamodel mapping coverage of subconstructs in functional linguistics to graphs}
     \label{fig:mappingCoverage}
\end{figure}

Table~\ref{tab:metamodelMappings} summarises the inter- and intra-mappings between the two metamodels of linguistics and graphs. Figure~\ref{fig:compareTax} represents a taxonomy of both metamodels and the mapping between the two presented domains. In Figure~\ref{subfig:mentalMapping} shows the mental (abstract) constructs contained by a machine. In contrast Figure~\ref{subfig:realisedMapping} show a concrete (literal) scenario in each domain. The bottom of the taxonomy in Figure~\ref{subfig:realisedMapping} represents concrete ideas in small units. As one moves up the respective taxonomies, it is observed how these smaller units can be combined to form more complex abstract representations. Lastly, Figure~\ref{fig:mappingCoverage} visually represents the coverage of which subcomponents from Functional Language Theory, could be mapped onto corresponding concepts in graph theory. Figure~\ref{subfig:presentMapping} represents the \textit{presence} of mappings between the subconstructs of the Functional Language Theory and graphs, while Figure~\ref{subfig:missingMapping} represents the absence of aforementiond mappings. That is the mental and realised metamodels of syntax, morphology and phonology could be mapped onto corresponding concepts in graphs. In contrast no mappings from the linguisitic mental and realised metamodels for pragmatic and semantics could be mapped onto graphs.

\begin{table}
\centering
\begin{tabular}{|p{3.7cm}|p{3.7cm}|}
\hline \hline
\multicolumn{2}{|c|}{\textbf{Linguistic Metamodels}}                           \\ \hline \hline
\multicolumn{1}{|c|}{\textbf{Mental}}                     & \multicolumn{1}{c|}{\textbf{Realised}} \\ \hline \hline
\multicolumn{2}{|c|}{\textbf{Phonology}}                            \\ \hline \hline
\multicolumn{1}{|l|}{Phone}      &                                  \\ \hline
\multicolumn{1}{|l|}{\tikzmark{la11}Phoneme\tikzmark{ra11}}    & Allophone                        \\ \hline \hline
\multicolumn{2}{|c|}{\textbf{\tikzmark{la12}Morphology\tikzmark{ra12}}}                           \\ \hline \hline
\multicolumn{1}{|l|}{\tikzmark{la13}Morpheme\tikzmark{ra13}}   & Allomorph                        \\ \hline \hline
\multicolumn{2}{|c|}{\textbf{Syntax}}                               \\ \hline \hline
\multicolumn{1}{|l|}{Syntax}     & Sentence                         \\ \hline
\end{tabular}
\begin{tikzpicture}
\draw[-{Triangle[width=18pt,length=8pt]}, line width=10pt](0,0) -- (1, 0);
\end{tikzpicture}
\begin{tabular}{|p{3.7cm}|p{3.7cm}|}
\hline \hline
\multicolumn{2}{|c|}{\textbf{Graph Metamodels}}                                                 \\ \hline \hline
\multicolumn{1}{|c|}{\textbf{Mental}}                     & \multicolumn{1}{c|}{\textbf{Realised}} \\ \hline \hline
\multicolumn{2}{|c|}{\textbf{Phonology}}                                              \\ \hline \hline
\multicolumn{1}{|l|}{Marking}             &                                           \\ \hline
\multicolumn{1}{|l|}{Lines \& Shapes}     & Lines and Shapes on medium                \\ \hline \hline
\multicolumn{2}{|c|}{\textbf{Morphology}}                                             \\ \hline \hline
\multicolumn{1}{|l|}{Entity/Relationship} & Vertex/Edge                               \\ \hline \hline
\multicolumn{2}{|c|}{\textbf{Syntax}}                                                 \\ \hline \hline
\multicolumn{1}{|l|}{Incidence Function}  & Graph                                     \\ \hline
\end{tabular}
\caption{Table representation of the inter-mapping between the metamodels within each domain as well as the intra-mapping between the corresponding metamodels across domains.}
\label{tab:metamodelMappings}
\end{table}

\section{Future work}
\label{sec:future}
Graphs structures exhibit many similar features as those found in a human language. Future work includes investigating whether deep structures exist in graph structures. If deep structures do exist, an analysis should be conducted, as to whether the deep structures in graphs can be mapped to the corresponding deep structures in linguistics. No mapping to the semantical and pragmatic concepts using Definition \ref{defn:graphBondyMurty} could be found. Further investigation into the introduction of semantic and pragmatic concepts into graphs should be undertaken. This includes whether the ideas can be mapped onto the corresponding subconstructs of linguistics. Additional analysis should be done to determine whether graphs could be classified as a linguistic theory, that is to say, whether graphs fulfill the elements of economy, simplicity, generality and falsifiability~\citep{HickeyLevelsLanguage}. In addition, any proposed linguistic theory should satisfy observational, descriptive and explanatory adequacy~\citep{Green2006LevelsExplanatory,Rizzi2016TheAdequacy,HickeyLevelsLanguage}.

\section{Conclusion}
\label{sec:conclusion}

From the investigation undertaken, there is a large overlap between concepts in linguistics and graphs. The subconstructs that appear missing from graph structures are the representation of semantic and pragmatic information. These missing subconstructs were identified by deriving mappings between two metamodels in both Fundamental Language Theory and mathematical graph structures. The metamodels considered are the mental and realised metamodels of each domain.

This reports highlights the missing concepts of semantics and pragmatics which occur when using mathematical graphs. Graphs lacking semantics and pragmatics could be viewed as being weak from the perspective of linguistics, that is lacking meaning. This paper proceeds to refer to mathematical structures void of semantics and pragmatics as \textit{etiolated structures}. In biological processes etiolation occurs in flowering plants that grow in partial or complete darkness. This process leads to elongated stems and leaves, longer internodes and chlorosis in plants~\citep{Armarego-Marriott2020BeyondStudies}. In an abstract context etiolation can be viewed as the removal of substance from an entity in question, thus the reason for naming mathematical structures that are void of semantics and pragmatics. An etiolated structure only contains syntactic information of the source structure and is often derived from another structure or representation. Examples of source structures include examples such as social networks or the relationship between cited authors. The original structure or representation is referred to as the \textit{source structure}, while the etiolated structure is referred to as the \textit{target structure}.

Formally the \textit{target structure} is said to be isomorphic to the \textit{source structure}, in other words preserving the structural information. For an in-depth discussion on isomorphisms, the reader is directed towards \textit{Graph Theory} by Reinhard Diestel \citep{Diestel2017GraphTheory}. The omission of semantic and pragmatic information eases the mathematical manipulation of the target structure. The target, acting as an isomorphic proxy to the source structure, is manipulated rather than the source structure itself. Isomorphisms allow human cognition to transcribe a problem and solution between two or more independent domains. The transcribing of the problem into a different domain might allow for easier solving of the problem. A transference of the solution back to the original domain may not be possible and needs to be researched further. \citep{Greer1998TheCognition,Uptegrove2004StudentsIsomorphisms}

\bibliographystyle{plainnat}
\bibliography{references}

\end{document}